\documentclass[10pt,conference]{IEEEtran}
\IEEEoverridecommandlockouts
\usepackage{cite}
\usepackage{amsmath,amssymb,amsfonts}
\usepackage{algorithmic}
\usepackage{graphicx}
\usepackage{textcomp}
\usepackage{xcolor}
\usepackage{comment}
\usepackage{tabularx}
\usepackage{multirow}

\def\BibTeX{{\rm B\kern-.05em{\sc i\kern-.025em b}\kern-.08em
    T\kern-.1667em\lower.7ex\hbox{E}\kern-.125emX}}
\IEEEpubid{Accepted for publication at GREENS 2025.}
\begin{document}

\title{Breaking the \textbf{ICE}: Exploring promises and challenges of benchmarks for \textbf{I}nference \textbf{C}arbon \& \textbf{E}nergy estimation for LLMs}

\author{\IEEEauthorblockN{Samarth Sikand$^\dagger$, Rohit Mehra$^\dagger$, Priyavanshi Pathania$^\dagger$, Nikhil Bamby$^\dagger$, Vibhu Saujanya Sharma$^\dagger$, \\Vikrant Kaulgud$^\dagger$, Sanjay Podder$^\ddagger$, Adam P. Burden*}
	\IEEEauthorblockA{\textit{$^\dagger$Accenture Labs, India \hspace{0.4em}
			$^\ddagger$Accenture, India \hspace{0.4em}
			*Accenture, USA}\\
		{\{s.sikand, rohit.a.mehra, priyavanshi.pathania, nikhil.bamby, vibhu.sharma, vikrant.kaulgud, sanjay.podder, adam.p.burden\}}\\@accenture.com}
}

\maketitle

\begin{abstract}
While Generative AI stands to be one of the fastest adopted technologies ever, studies have made evident that the usage of Large Language Models (LLMs) puts significant burden on energy grids and our  environment. It may prove a hindrance to the Sustainability goals of any organization. A crucial step in any Sustainability strategy is monitoring or estimating the energy consumption of various components. While there exist multiple tools for monitoring energy consumption, there is a dearth of tools/frameworks for estimating the consumption or carbon emissions. Current drawbacks of both monitoring and estimation tools include high input data points, intrusive nature, high error margin, etc. We posit that leveraging emerging LLM benchmarks and related data points can help overcome aforementioned challenges while balancing accuracy of the emission estimations. To that extent, we discuss the challenges of current approaches and present our evolving framework, \textit{R-ICE}, which estimates prompt level inference carbon emissions by leveraging existing state-of-the-art(SOTA) benchmark. This direction provides a more practical and non-intrusive way to enable emerging use-cases like dynamic LLM routing, carbon accounting, etc. Our promising validation results suggest that benchmark-based modelling holds great potential for inference emission estimation and warrants further exploration from the scientific community.	
\end{abstract}

\begin{IEEEkeywords}
LLM, Generative AI, Carbon Emissions, Inference, Sustainability
\end{IEEEkeywords}

\section{Introduction}\label{sec:intro}
Today Generative AI has been recognized as one of the most disruptive technologies in the last few decades and is responsible for driving transformation across all industries at an incredible pace. As a result, a burgeoning marketplace for LLM-powered tools and applications has emerged and is thriving; for example, popular LLM-powered tools like ChatGPT and Github Copilot are serving millions of customers daily.\cite{chien2023reducing,githubcopilot}.

\begin{comment}
Today, Generative AI has become a core component of almost every organization’s innovation and transformation agenda. LLM-powered tools, like ChatGPT, are serving 100 million users monthly and handling more than 10 million queries per day \cite{chien2023reducing}. . Currently, there exists a burgeoning marketplace for LLM-powered tools and applications. Among its myriad of applications and uses spanning industries/domains, software engineering stands out the most for being rapidly and radically transformed. Organizations are trying to streamline and optimize their Software development process by relying heavily on Generative AI. For example, Github Copilot\cite{githubcopilot} is being used by millions of developers every day to accelerate their everyday coding tasks
\end{comment}

While LLMs have demonstrated tremendous reasoning and deduction capabilities, they come at the expense of massive compute utilization and environmental costs. Multiple studies\cite{patterson2022carbon,schwartz2020green} have highlighted the massive carbon footprint associated with training as well as inferencing LLMs. In Google’s recent environmental report, they reported a 48\% increase in their total GHG(Greenhouse Gas) emissions between 2019 and 2023\cite{google24env}, attributing the increased energy demand to AI being baked into their products. While the carbon cost of training Gen AI models is massive, it is quickly eclipsed by the carbon costs of inferencing, given there are millions or billions of inference requests served per day. Tracking such carbon emissions can become a serious impediment to any organization's Sustainability journey.

A crucial component of any Sustainability strategy is to monitor or estimate the energy usage of their processes/products. Currently, there exists a plethora of energy monitoring tools, like CodeCarbon\cite{benoit_courty_2024_11171501} and Eco2AI\cite{budennyy2022eco2ai}, that can monitor energy consumption from certain hardware components. Moreover, there exists few energy and carbon estimation tools, like LLMCarbon\cite{faiz2024llmcarbon}, which rely on LLM architectural details to perform its analytical modelling.
However, current approaches suffer from several shortcomings, including high error of margin, intrusiveness, high input parameters. Moreover, current trend reveals that most organizations prefer relying on LLM API services from vendors, like OpenAI, to integrate GenAI into their workflow and products. A major drawback of using such services is that the vendors rarely disclose the LLM technical details and the infrastructure they are deployed on.
\IEEEpubidadjcol
The aforementioned drawbacks restrict stakeholders from incorporating carbon and energy costs into key design decisions associated with LLM management and routing. To tackle the challenges, we shift our focus towards the emerging landscape of LLM benchmarks related to efficiency and task performance. The benchmark results will allow us to perform certain modelling of prompt-level inference energy and cost estimation. To that extent, we put forth our evolving framework, \textit{Regression-based Inference Carbon and Energy estimation (R-ICE)}, for accurate estimation of carbon and energy costs at a prompt-level granularity in pre-inferencing scenarios.
\begin{comment}
	\begin{itemize}
		\item The monitoring tools are intrusive, i.e., they need to be set up within the LLM deployment environment. Access to such environments (e.g. OpenAI's GPT-4 or GPT-3.5 environments) may be restricted or challenging. Additionally, these tools work in post-inferencing scenarios, i.e., when inferencing is complete.
		
		\item The analytical models tend to be reliant on model architecture details. Such models may not work for LLMs with restrictive/commercial licenses where architecture details are not released(e.g., GPT-4). Moreover, such models may rely on Floating Point Operations(FLOPs) computation to compute metrics like latency, but few studies\cite{narayanan2023cheaply} have shown them to be quite inaccurate.
		
		\item The analytical models may overgeneralize certain assumptions, which can lead to huge error margins in their estimates.
		
		\item The monitoring tools, analytical models, and CSP’s Carbon	calculators may not provide emission values at prompt-level	granularity.
		
	\end{itemize}
\end{comment}

Our early validation results show that our framework achieves an average prediction error of  $\sim$15\%, while mitigating many of the mentioned challenges. The results highlight promising potential of leveraging benchmarks and underscore the need for such practical and non-intrusive approaches which can easily be adopted and scaled at an enterprise level.

\begin{comment}
	To mitigate the aforementioned challenges, we propose a novel data-driven framework, RICE, for accurately estimating carbon and energy costs at a prompt-level granularity in \textit{pre-inferencing} scenarios. The recent rise of benchmarks related to performance and efficiency allows us to perform certain modelling of prompt-level inference energy and carbon cost estimation. Holistic Evaluation of Language Models (HELM)\cite{liang2022holistic} is one such popular benchmark for evaluating State-of-the-Art (SOTA) LLMs on various metrics. An extension of the HELM benchmark evaluates SOTA LLMs on efficiency metrics, like Latency, as well. For our work, we exclusively use the empirical data points put forth by their research. The reasons for exclusively choosing HELM-efficiency include the open-source license of their research data and the wide spectrum of LLM model sizes covered in their study. In subsequent sections, we briefly describe the HELM benchmark, delve deep into our estimation framework, validate our approach, highlight a few shortcomings, and finally discuss future work.
\end{comment}

\section{Background and Motivation}
In this section, we delve deeper into the existing monitoring/estimation approaches and highlight the challenges associated with them in detail. Further, we explain why our focus shifted towards finding more practical estimation approaches.

\subsection{Existing Approaches}\label{subsec:approaches}

\paragraph{Energy Monitoring Tools}

In our survey, we assessed more than 20 tools, like eco2AI\cite{budennyy2022eco2ai}, where each tool differentiated itself by monitoring all or different subsets of underlying hardware (e.g., GPU, RAM, CPU, storage) for every running workload. They also differed in the kind of support provided for a wide range of hardware-specific models, like Intel or AMD CPUs. Among them, CodeCarbon \cite{benoit_courty_2024_11171501} emerged as one of the more popular tools used across multiple studies due to its ease of use and granular monitoring capabilities.

\paragraph{Estimation Tools and Frameworks}
Most estimation approaches we surveyed developed an analytical model to estimate energy consumption and carbon emissions. Such techniques rely heavily on the model architecture details to compute FLOPS (Floating Point Operations), which in turn is leveraged to compute energy cost. LLMCarbon\cite{faiz2024llmcarbon}, one such example, proposed an end-to-end methodology for estimating the carbon footprint of LLMs during training and inferencing. Other estimation tools, like MLCO2 Impact\cite{lacoste2019quantifying}, provide a rough carbon emission estimate at a broad level based on a few parameters.  

\paragraph{Cloud Service Provider(CSP) calculators}
CSPs, like Amazon Web Services, also provide their own tools\cite{aws_ccft} for estimating carbon emissions of the workloads running in their respective cloud instances.

\subsection{Challenges}
Analysis of the existing approaches(Section \ref{subsec:approaches}) have highlighted certain shortcomings:
\begin{itemize}
	\item \textbf{Intrusiveness}: The monitoring tools are intrusive, i.e., they need to be
	set up within the LLM deployment environment. Access to such environments may be restricted or challenging. Additionally, these tools work in post-inferencing scenarios, i.e., when inferencing is complete.
	\item \textbf{Restrictive licenses of LLMs} : The analytical models tend to be reliant on model architecture 	details. Such models may not work for LLMs with restrictive licenses where architecture details are not released(e.g., GPT-4). Moreover, such models may rely on Floating Point Operations(FLOPs) computation to compute metrics like latency, but few
	studies [13] have shown them to be quite inaccurate.
	\item \textbf{Over-Generalization}: The analytical models may overgeneralize certain assumptions, which can lead to huge error margins in their estimates.
	\item \textbf{Granularity}: The monitoring tools, analytical models, and CSP’s
	Carbon calculators may not provide emission values at prompt-level granularity.
\end{itemize}

\subsection{Motivation : Towards a more Practical Approach}
While analyzing the existing approaches in an enterprise setting, the challenges mentioned in the previous sub-section become a serious bottleneck in scaling them. We conducted broad discussions with various stakeholders within our organization, regarding use of such tools, and noted that their concerns focused on out-of-the-box estimation capability with minimal input data points and acceptable level of accuracy/error margin. Current approaches fail to balance their expectations and impede their adoption at an enterprise level.

Based on the feedback, we began exploring alternative avenues and recognized the underutilized potential of LLM benchmarks.
\begin{comment}
The landscape LLM benchmarks is rapidly progressing and evolving as it covers various aspects of LLMs like task comprehension, ethics, robustness, efficiency etc. Moreover, most benchmarks and associated results are publicly available, it allows to model efficiency behavior, wherever possible, even for LLMs with restrictive licenses. While we leverage efficiency(e.g., latency) benchmarks, and related datasets, of LLMs to demonstrate our view in this work, there is great potential for community to explore and build in this direction.
\end{comment}

\section{Holistic Evaluation of Language Models(HELM)} \label{sec:helm}
HELM\cite{liang2022holistic} is a popular framework that evaluates the performance of SOTA LLMs across popular benchmarks of logical reasoning, question answering, mathematical reasoning, etc. and provides a common platform that contrasts and compares different LLMs on specified metrics.

\begin{figure*}[ht]
	\centerline{\includegraphics[width=0.8\textwidth]{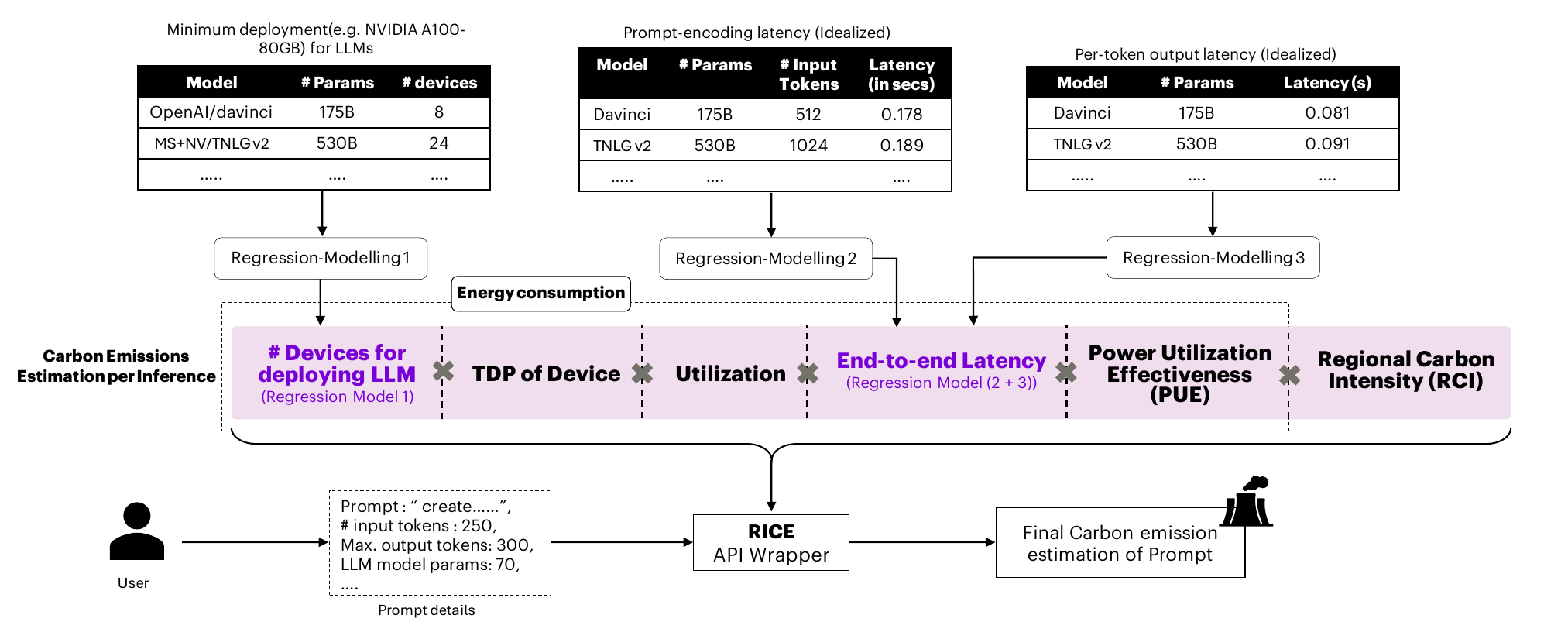}}
	\caption{Overall approach to RICE.}
	\label{fig:rice_struc}
\end{figure*}

\subsection{Inference Efficiency metrics}\label{sec:inf_eff}
HELM authors further extended their work by creating a cost model for Auto-regressive transformer models, which are the basis of most LLMs\cite{narayanan2023cheaply}. They propose that end-to-end runtime, for any given prompt or query, can be broken down into two parts: piecewise linear function of input prompt tokens and linear function of generated output tokens. Eq. 1 demonstrates the breakdown of the end-to-end runtime($T$) for any given prompt.

\begin{equation}\label{eq:1}
	T = prompt\_encoding\_time(p)+(o-1)\beta
\end{equation} 

$p$ and $o $ denote the number of input prompt and output tokens, respectively. $\beta$ denotes additional runtime to generate an output token. The authors propose two versions of the runtime metric $T$, namely \textit{\textbf{Idealized runtime}} and \textit{\textbf{Denoised runtime}}.

\begin{comment}
	\textit{\textbf{Idealized runtime}} represents the end-to-end runtime of an inference request on a particular LLM architecture as well as a known hardware and software stack(e.g., NVIDIA A100 GPU and MegatronLM). Whereas, \textbf{\textit{Denoised runtime}} represents best-case runtime for a prompt when using commercial black-box models via APIs, after factoring out noise due to performance variations. Given certain assumptions that underlie our framework (further in Section \ref{sec:data_assump}), we leverage the notion of \textit{Idealized runtime} instead of \textit{Denoised runtime}. Limited or absent technical data around model architecture and infrastructure makes it challenging to use Denoised runtime.
\end{comment}
 
\textit{Idealized runtime} is based on the notion that LLMs are deployed on standardized hardware and software stack, requiring only a minimum number of hardware units. It also allows for a uniform comparison w.r.t carbon efficiency amongst various LLMs. Hence, in our approach, Runtime is assumed to be \textit{Idealized} runtime.

\section{Regression-based Inference Carbon and Efficiency estimations (RICE)}\label{sec:rice}

The overall carbon footprint of an AI model can essentially be a sum of two parts: \textit{Embodied carbon emissions} and \textit{Operational carbon emission}\cite{wu2024beyond}. In our framework, we focus only on estimating \textit{Operational} emissions of LLMs. A standard formulation of such emissions, assuming the workload running on the Cloud, is based on certain factors: the number of units for each hardware type, TDP value of each hardware type, hardware utilization during processing, Runtime of process, Power Utilization Efficiency(PUE) of Data Center and Regional Carbon Intensity(RCI)

\begin{equation*}\label{eq:22}
	\begin{split}
		CO_2 eq = \sum_{i=0}^{num\_hardware}&(\:\# \; hardware_i \times Power_i \\
		& \times runtime \times PUE \times RCI)
	\end{split}	
\end{equation*}

Fig.\ref{fig:rice_struc} represents our overall approach to Regression-based Inference Carbon \& Efficiency estimation(RICE). A high-level overview of the various components are detailed in Table \ref{tab:high_level}

In our approach we create \textit{three} regression models : (i) First, to estimate the minimum number of devices to load a LLM model, (ii) Two regression models to estimate end-to-end latency for any given input. In further sub-sections, we discuss the certain assumptions underlying our approach, Dataset curation, Regression modelling for E2E latency and min. hardware units. 

\subsection{Dataset Curation and Assumptions}\label{sec:data_assump}

In the following sub-sections: (i) we delineate few of the assumptions made to make our framework accurate and precise and, (ii) describe the process of curating data points from HELM efficiency benchmark.

\begin{comment}
	As mentioned in Section \ref{sec:intro}, LLMs with varying licenses(commercial, open-source, etc.) pose certain challenges for estimating inference time emissions by obscuring crucial technical and infrastructure details. Section \ref{sec:assumptions} delineates a few of the assumptions made to make our framework accurate and precise. Further, Section \ref{sec:dataset_curation} explores the details of how data points from the HELM efficiency benchmark are leveraged.
\end{comment}

\begin{table}[t]
	\centering
	\tiny
	\caption{High-level description of components in Fig.\ref{fig:rice_struc}}
	\begin{tabularx}{\columnwidth}{|c|>{\centering\arraybackslash}X|}
		\hline
		&\\[-1em]
		\textbf{Component}& \textbf{Description}\\
		\hline
		&\\[-1em]
		\# of Devices & Min. number of hardware units to deploy a LLM \\ \hline
		&\\[-1em]
		Thermal Design Power (TDP) & Power consumption of hardware under maximum theoretical load \\ \hline
		&\\[-1em]
		Utilization & Average Hardware utilization/efficiency \\ \hline
		&\\[-1em]
		End-to-end(E2E) latency & Total time taken to process the input prompt and generate the final response(till the last output token) \\ \hline
		&\\[-1em]
		Power Utilization Effectiveness (PUE) & metric to determine the efficiency of Data Centers(DC) \\ \hline
		&\\[-1em]
		Regional Carbon Intensity (RCI) & Amount of Carbon dioxide equivalent($CO_2 eq$) emitted per kilowatt-hour (KWh) of electricity consumed \\ \hline
	\end{tabularx}
	\label{tab:high_level}
\end{table}

\subsubsection{Assumptions}\label{sec:assumptions}

As our approach relies heavily on the data points collated from HELM-Efficiency\cite{narayanan2023cheaply}, we inherit some of the assumptions from their experimental setup as well. Few of the assumptions are listed as follows:

\begin{itemize}
	\item LLMs are deployed on a minimum number of accelerator units of a single hardware model(NVIDIA A100-80GB)
	
	\item The default Hardware Utilization value is assumed to be 0.26 (or 26\%). The assumption are based on a study that estimated the footprint of the BLOOM model\cite{luccioni2023estimating}. Similarly, the default value of PUE is assumed to be 1.1. Our approach allow custom/real-time values for PUE and Utilization to be plugged in, if required.
	
	\item The framework relies on the most fundamental feature of LLM, i.e. Number of Parameters to be given to our framework.
	
	\item The E2E latency formulation(Eq.\ref{eq:1}) is applicable to context windows with $<10k$ tokens.
	
	\item The batch size for inference is considered to be 1. We consider batch size as 1 for our modelling and evaluation as well.
\end{itemize}

\subsubsection{Dataset Curation}\label{sec:dataset_curation}
HELM-Efficiency\cite{narayanan2023cheaply} authors were provided access to popular black-box models, like OpenAI GPT-3, Anthropic Claude, AI21 Jurrasic model, etc, which were then deployed on a standardized hardware stack of NVIDIA A100 GPUs.
Less than 15 data points were available to model the minimum number of devices required for deployment. Table \ref{tab:min_hardware_1} showcases a few of the sample data points for Min. number of devices.

\begin{table}[h]
	\centering
	\caption{Sample Data points for Minimum Hardware Deployment. Hardware or GPU-type for every LLM : NVIDIA A100-80GB}
	\begin{tabularx}{\columnwidth}{c>{\centering\arraybackslash}X>{\centering\arraybackslash}X}
		\hline
		&&\\[-1em]
		\textbf{Model Name} & \textbf{\# Params (B)} & \textbf{\# GPUs}\\
		\hline
		&&\\[-1em]
		EleutherAI/GPT-J & 6 & 1 \\ \hline
		&&\\[-1em]
		AI21/J1-Grande v1 & 17 & 1 \\ \hline
		&&\\[-1em]
		Anthropic/v4-s3 & 52 & 4 \\ \hline
	\end{tabularx}
	\label{tab:min_hardware_1}
\end{table}

For modelling E2E latency, the authors leveraged Eq.\ref{eq:1} which breaks down the final latency into \textit{two} components: \textit{Prompt Encoding latency} and \textit{Per-output token latency}. They ran extensive experiments with varying input prompt lengths, ranging from 1 to 1920 tokens, and captured the prompt encoding time for the varying prompt lengths. Similarly, by varying the output token lengths, ranging between 1 and 64, the authors measured the average per-output token latency for each LLM.
\begin{comment}
	Table 1 and Table 2 showcase some of the empirical data points collated for Prompt encoding and per-output token latency, respectively.
\end{comment}

\begin{figure}[h]
	\centering
	\includegraphics[width=\linewidth]{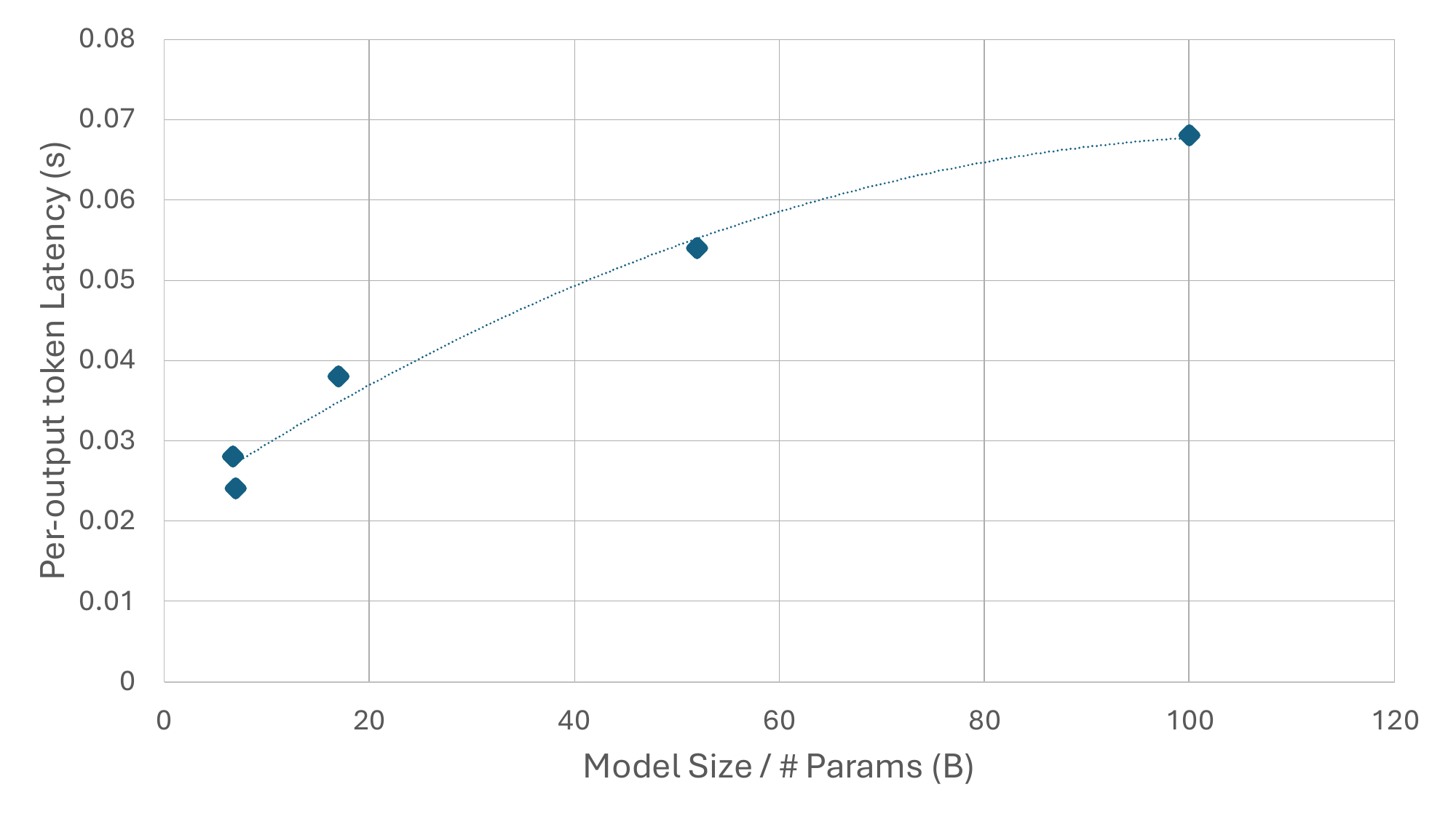}
	\caption{Correlation Per-output Token Latency and Model Size(\# Parameters) of sample data points}
	\label{fig:per_output_token}
\end{figure}

Prompt-encoding latency dataset consists of $\sim$400 data points whereas the Per-output token latency dataset was restricted to 10 data points. Fig.\ref{fig:per_output_token} highlights the correlation between LLM size and Per-output token latency for a few of the data points.

\subsection{Regression Modelling}

Post analyzing all the data points, we observe a linear correlation between Minimum deployment hardware and LLM size(\# parameters). Based on the observation, we train a Linear Regression model on all available data points to compute the best-fit line. The \textit{Coefficient of Determination}($R^2$) was observed to be close to $1.0 (0.99)$ and concluded that minimum deployment hardware shows a linear relationship with Model Size.

Table \ref{tab:regression_algos} details the various algorithms used to model Prompt-encoding latency on a standard train-test(80:20) split. To evaluate the modelling performance of various algorithms, the regression metrics like \textit{Mean Absolute Percentage Error(MAPE)}, \textit{Mean Squared Error(MSE)} and \textit{Coefficient of Determination($R^2$)} were utilized. For estimating prompt encoding latency in the final Carbon emission formulation(see Fig. \ref{fig:rice_struc}), the Random Forest model is leveraged.

Similar to Min. deployment hardware, the data points for per-output token latency were fairly low. We train a Linear and Polynomial Regression model to find the best-fit curve for all the observed data points. The Polynomial Regression model is observed to give the best-fit curve with a $R^2$ value of $0.94$.

Finally, for any given prompt, the Random Forest (Regression Model 2) and Polynomial Regression (Regression Model 3) models are utilized to estimate \textit{Prompt-encoding} and \textit{Per-output token} latency, respectively. The values are substituted in Eq.\ref{eq:1} to calculate the final E2E latency (refer Fig.\ref{fig:rice_struc}). Similarly, values from the Linear Regression model (Regression Model 1) for estimating \textit{Min. \# Devices} are substituted as well.

The rest of the components of our approach (Fig.\ref{fig:rice_struc}) provides flexibility to incorporate values using different modes, e.g., default values, querying using a third-party API or user inputs.

\begin{table}[t]
	\centering
	\tiny
	\caption{Regression Algorithms evaluated for Prompt Encoding Latency. Lower \textit{MAPE} is better ($\downarrow$)}
	\begin{tabularx}{\columnwidth}{c|>{\centering\arraybackslash}X>{\centering\arraybackslash}X>{\centering\arraybackslash}X}
		\hline
		&&&\\[-1em]
		\textbf{Algorithm} & \textbf{$R^2$} & \textbf{MSE} & \textbf{MAPE} $\downarrow$ \\
		\hline
		&&&\\[-1em]
		Random Forest & 0.986 & 0.001 & 0.050 \\ \hline
		&&&\\[-1em]
		XGBoost & 0.976 & 0.003 & 0.60 \\ \hline
		&&&\\[-1em]
		Gradient Boosting & 0.946 & 0.007 & 0.070 \\ \hline
		&&&\\[-1em]
		Decision Tree & 0.965 & 0.004 & 0.084 \\ \hline
		&&&\\[-1em]
		SVR (RBF kernel) & 0.905 & 0.013 & 0.171 \\ \hline
		&&&\\[-1em]
		SVR (Linear) & 0.712 & 0.040 & 0.507 \\ \hline
		&&&\\[-1em]
		Elastic Net & 0.743 & 0.036 & 0.594 \\ \hline
		&&&\\[-1em]
		SGD & 0.741 & 0.036 & 0.597 \\ \hline
		&&&\\[-1em]
		Linear Reg. & 0.747 & 0.035 & 0.608 \\
		\hline
	\end{tabularx}
	\label{tab:regression_algos}
\end{table}

\section{Early Validation and Discussion}\label{sec:eval}

To validate the end-to-end approach, we leverage an external efficiency benchmark dataset, namely HuggingFace \textbf{\textit{LLM-Perf Leaderboard}\cite{llm-perf-leaderboard}}, whose data points are unseen by our developed regression models. LLM-Perf Leaderboard evaluates various open-source models, which are part of Open-LLM leaderboard, on various efficiency metrics, including prefill latency, E2E latency ($E2E$), tokens per kWh of energy($tok/kWh$). 

While LLM-Perf Leaderboard contains more than 500 measurement points, we shortlisted only 46 data points for our evaluation because there are certain nuances and challenges in the Leaderboard measurements. LLMPerf leaderboard contains multiple variations of the same LLM based on various optimization permutations, like Quantization\cite{frantar2022gptq}, Flash Attention\cite{dao2022flashattention} etc., which leads to duplicate entries with different energy and latency values if we omit the optimization features. Hence, we apply certain criteria that filter out data points where any optimization is applied. We attempt to maximize data points for our evaluation by setting the aforementioned filter criteria and also widening the spectrum of model sizes. 

To ensure uniformity in evaluation, we set the evaluation prompt input tokens at 192 and max. output tokens at 250. The ground-truth values for metrics under test are scaled for the evaluation prompt using Eq.\ref{eq:2} and MAPE is computed along with estimated metric values(Fig.\ref{fig:rice_struc})

\begin{comment}
To ensure uniformity in evaluation, we estimate the E2E latency and Energy consumed($Est.\:Metric$) via our approach(see Fig.\ref{fig:rice_struc}) and compare it with Adjusted Empirical Values($AEV$) computed from Leaderboard data points for our evaluation prompt. The evaluation prompt has 192 input tokens and Max. Output tokens were set to 250. Eq.\ref{eq:2} demonstrates how AEV, for energy and latency, is computed for the evaluation prompt. We leverage \textit{Mean Absolute Percentage Error (MAPE)} to evaluate the performance of our approach. $n$ denotes the 46 data points filtered from Leaderboard.
\end{comment}

\begin{equation}\label{eq:2}
	\begin{split}
		&Scaled(E2E\:Lat.) = \frac{E2E}{256}*(Output\:tokens)\\
		&Scaled(Energy) = \frac{Output\:tokens}{tok/kWh}\\
	\end{split}
\end{equation}

The MAPE values for E2E latency and Energy consumption metrics, as shown in Table \ref{tab:validation}, demonstrate the promising results for our approach. The average MAPE values for Latency and Energy are 22.9\% and 15.3\%, respectively. The results indicate that our approach can estimate energy consumption, and in turn carbon emissions, accurately at a prompt level. Table \ref{tab:sample_preds} showcases estimated latency and estimated metrics for a few data points from LLM-Perf-Leaderboard.

\section{Limitations}

\begin{table*}[t]
	\centering
	\begin{minipage}{.3\textwidth}
		\centering
		\caption{Early Validation Results}
		\label{tab:validation}
		\begin{tabular}{c|c}
			\hline
			&\\[-1em]
			\textbf{Target Metric} & \textbf{MAPE}\\ \hline
			&\\[-1em]
			E2E Latency & 22.9\% \\
			Energy consumption & 15.3\%\\ \hline
		\end{tabular}
	\end{minipage}\hfill
	\begin{minipage}{0.7\textwidth}
		\tiny
		\centering
		\caption{Sample predictions on select External LLM-Perf Leaderboard data points}
		\label{tab:sample_preds}
		\begin{tabular}{c|c|c|c||c|c}
			\hline
			\textbf{Model Name} & \textbf{\# Params(B)} & \textbf{E2E latency(s)} & \textbf{Predicted Latency(s)} & \textbf{Energy (J)} & \textbf{Predicted Energy (J)}\\
			\hline
			&&&&&\\[-1em]
			cerebras/cerebras-gpt-1.3b & 1.30 & 3.75 & 3.41 & 3.99 & 3.99 \\
			eleutherai/gpt-neo-2.7b & 2.72 & 5.76 & 4.42 & 6.12 & 5.17 \\
			facebook/xglm-4.5b & 6.99 & 5.08 & 5.51 & 7.81 & 6.77 \\
			eleutherai/pythia-12b & 12.00 & 7.21 & 7.46 & 10.12 & 11.51 \\
			eleutherai/gpt-neox-20b & 20.74 & 10.30 & 9.05 & 15.77 & 15.24 \\ \hline
		\end{tabular}
	\end{minipage}
\end{table*}

Our initial validation results indicate that our framework can provide accurate, robust and consistent estimates at an individual prompt level. However, we acknowledge there exist certain limitations to our approach which can lead to inaccurate estimates.
\begin{itemize}
	\item Datasets size to train regression models esp. \# Devices and Per-output token is small.
	\item With HELM-efficiency benchmark as an exception, most LLM benchmarks, related to efficiency, deploy LLM on a single hardware unit for their experiments. This puts a restriction on the model size which can be deployed.
	\item Popular LLM services serves millions of requests daily, due to which they may adopt optimal and efficient deployment strategies like batching. Our assumption of singleton batch inference may not hold.
	\item Today most LLM services have a minimum context length of 100k tokens, which is much larger than our assumption of 10k tokens.
	\item Custom/proprietary software and hardware stack employed by vendors can significantly vary energy efficiency.
\end{itemize}

\section{Potential Impact on Software Industry}
Among all the industry domains, Software Engineering’s GenAI transformation outpaces every other domain. But integration of Gen AI services in Software Development lifecycle (SDLC) processes may come at an environmental cost. During the design stages of SDLC, architects and other relevant stakeholders have a broad view of the various factors that need to be considered, but often the carbon cost factor is absent from their view. Approaches, akin to R-ICE, allow us to bring in aspects of carbon and energy costs during the design stages. With an abundance of LLMs present in the market, it is important to understand the emissions associated with different LLMs so as to make an informed choice. To this extent, \textit{R-ICE}-like frameworks can help create a Green Ranking of LLMs, which can be plugged into various LLM orchestration and design engines. The purpose of such engines is to optimize for accuracy and latency, among other factors.

We believe that in software services organizations, the focus is more on GenAI applications and usage, which makes our framework an extremely valuable tool for all stakeholders. Reducing even a fraction of $CO_2 eq$ from each inference can vastly improve any organization's Sustainability quotient.

\section{Future Work and Conclusion}

In this paper, we have highlighted some of the challenges associated with estimating carbon emissions, energy consumption, and other efficiency metrics in the current landscape of tools. Our evolving work, R-ICE, presents a pragmatic, non-intrusive and precise first step in mitigating many challenges associated with current approaches. It also showcases the immense potential of LLM benchmarks in tackling estimation issues and hopefully motivates our community to explore this research direction further. To mitigate some of the shortcomings of the work, we are working towards leveraging some popular and upcoming efficiency benchmarks to create a hybrid approach that would make the framework more robust and take into account various optimizations associated with inferencing.

\bibliographystyle{IEEEtran}
\bibliography{IEEEabrv,IEEEexample}

\end{document}